\documentclass[conference]{IEEEtran}
\usepackage{times}

\usepackage[numbers]{natbib}
\usepackage{multicol}
\usepackage{graphicx}
\usepackage{booktabs}
\usepackage{amsmath,amssymb}
\usepackage{tikz}
\usetikzlibrary{positioning, arrows.meta}
\usepackage[bookmarks=true]{hyperref}
\graphicspath{{./}{figures/}}

\begin{document}

\title{\LARGE Topology-Agnostic Mesh Reconstruction of Deformable\\Objects from Sparse Touch}

\author{\authorblockN{\large Everest Yang}
\authorblockA{\large Brown University\\
everest\_yang@brown.edu}}


\maketitle
\thispagestyle{empty}
\pagestyle{empty}

\begin{abstract}
Estimating the full shape of a deformable object is especially challenging when vision is
unavailable: in the dark, inside an opaque bag, behind the manipulating hand, or under heavy self-occlusion. Touch is the natural sensor in these settings, but touches are sparse and local. We present a single \emph{topology-agnostic} estimator that reconstructs the full mesh of a deformable object from only a few
touches and no vision, using one permutation-invariant cross-attention architecture that handles a 1D rope, a 2D cloth, and a 3D volumetric soft body. The learned estimator reduces reconstruction error by roughly two-thirds relative to non-learned geometric mesh completion and a Gaussian-process surface baseline, and it outperforms a simpler global-pool set encoder, with the gap growing as more touches are observed. We then show that the estimator's deep-ensemble uncertainty can be used to learn where to touch next, which lowers error further and beats both random touching and a Gaussian-process active baseline at sparse budgets. This gain is modest on average but grows with self-occlusion and on the error tail. When vision is also available, where to touch barely matters, motivating the vision-free setting we study.
\end{abstract}

\IEEEpeerreviewmaketitle

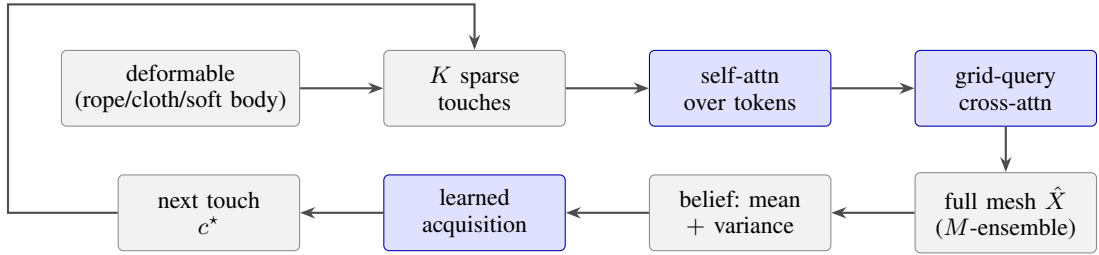
\begin{figure*}[t]
\centering
\begin{tikzpicture}[
    font=\small, node distance=6mm and 11mm,
    box/.style={draw, rounded corners=2pt, align=center, inner sep=5pt,
                minimum height=10mm, minimum width=24mm},
    accent/.style={box, fill=blue!12, draw=blue!65!black},
    plain/.style={box, fill=black!5, draw=black!45},
    arr/.style={-{Stealth[length=2mm]}, thick, black!70},
]
\node[plain] (obj) {deformable\\ (rope/cloth/soft body)};
\node[plain, right=of obj] (touch) {$K$ sparse\\ touches};
\node[accent, right=of touch] (enc) {self-attn\\ over tokens};
\node[accent, right=of enc] (dec) {grid-query\\ cross-attn};
\node[plain, below=of dec] (mesh) {full mesh $\hat{X}$\\ ($M$-ensemble)};
\node[plain, left=of mesh] (bel) {belief: mean\\ $+$ variance};
\node[accent, left=of bel] (acq) {learned\\ acquisition};
\node[plain, left=of acq] (next) {next touch\\ $c^\star$};
\draw[arr] (obj) -- (touch); \draw[arr] (touch) -- (enc); \draw[arr] (enc) -- (dec);
\draw[arr] (dec) -- (mesh); \draw[arr] (mesh) -- (bel); \draw[arr] (bel) -- (acq);
\draw[arr] (acq) -- (next);
\draw[arr] (next.west) -| ([xshift=-7mm]obj.west) |- ([yshift=6mm]obj.north) -| (touch.north);
\end{tikzpicture}
\caption{One topology-agnostic estimator reconstructs a 1D, 2D, or 3D deformable mesh from sparse touches via grid-query cross-attention. A deep ensemble gives a per-vertex belief and a learned acquisition picks the next touch (blue: learned).}
\label{fig:pipeline}
\end{figure*}

\section{Introduction}
Manipulating a deformable object, whether cloth, rope, or cable, starts with a hard
perception problem: its configuration is effectively infinite-dimensional, so a robot
must estimate the object's current shape before it can fold, route, or grasp it.
State estimation is thus a prerequisite for almost every deformable-manipulation task.

Almost all existing deformable perception assumes vision \cite{garmentnets, meshoccl, diffdyn}. Yet the settings where deformable manipulation matters most are often those where vision is degraded or unavailable. Consider reaching into a bag or drawer, operating in darkness, manipulating an object in-hand where the gripper hides it, or handling heavily crumpled cloth whose folded-under layers no external camera can see. Touch is then the natural and often the only
reliable sensor. But each contact is local and costly, so a practical system can
afford only a handful of them. This raises two coupled questions. First, can a robot reconstruct the \emph{full}
shape from a handful of sparse, noisy touches and no vision? Second, can it decide
\emph{where to touch next} to learn the most per touch? We study both and separate them: estimation first, active sensing on top.

Our \textbf{primary contribution} is a single \emph{topology-agnostic} estimator
that reconstructs the full vertex mesh from sparse touch alone. It is a
permutation-invariant cross-attention network \cite{vaswani2017attention,
zaheer2017deepsets}: touch observations are decoded via a fixed bank of
grid-position queries, one per output vertex, so changing only the query-grid
dimensionality yields the same model for a 1D rope, 2D cloth, or 3D soft body.
Prior active-touch work targets rigid objects \cite{smith21} or vision-led cable
splines \cite{mazza26}; we are not aware of a tactile-only full-mesh estimator
demonstrated across all three. A deep ensemble \cite{lakshminarayanan2017ensembles}
provides per-vertex uncertainty.

Our \textbf{secondary contribution} is learned active sensing: an acquisition
policy trained against an expected-error-reduction oracle uses ensemble uncertainty
to pick the next touch. It beats random touching and a Gaussian-process active
baseline \cite{yi16} at sparse budgets, though the average gain is modest and grows
with self-occlusion and on the error tail. When vision is also available, vision and
the learned prior together pin down the hidden state and touch placement barely
matters, motivating the vision-free setting we study.

In summary, we contribute: (i) a topology-agnostic cross-attention estimator for
full-mesh reconstruction from sparse touch across 1D, 2D, and 3D deformables;
(ii) a learned active policy with occlusion-dependent gains at sparse budgets; and
(iii) a negative result delimiting when active placement helps.

\section{Related Work}

\textbf{Deformable estimation from vision:}
Most deformable shape estimation recovers full state passively from vision: garment-mesh completion from depth \cite{garmentnets}, cloth-mesh reconstruction with occlusion reasoning \cite{meshoccl}, diffusion models that
generate full cloth states for model-predictive control \cite{diffdyn}, and rope
state from a single occluded point cloud \cite{lvdlo}. Separate lines of work learn
dynamics across topologies \cite{gdoom, dpinet}, but assume near-full visual state
and predict dynamics rather than reconstructing a mesh from sparse touch. These
methods are vision-led and passive. We
instead reconstruct from touch alone, actively select observations, and use one
model across 1D, 2D, and 3D objects.

\textbf{Active tactile shape estimation:}
Choosing where to touch to recover shape has been studied mainly for \emph{rigid}
objects: learned next-best-touch from vision and touch on CAD models \cite{smith21},
classical GP uncertainty sampling \cite{yi16, gpis}, and active visuo-haptic completion
that contacts points of greatest ensemble disagreement \cite{actvh}, in the spirit
of expected-error reduction \cite{roy2001eer}. Visuo-tactile methods such as
NeuralFeels \cite{neuralfeels} reconstruct in-hand shape passively with vision. On
deformables, Mazza \emph{et al.} \cite{mazza26} probe visually occluded cable
segments, but are vision-led, use a geometric heuristic rather than a learned
policy and output a 1D spline rather than a mesh. Act-VH's disagreement heuristic is the closest acquisition precedent; we show the analogous variance rule degrades below random on deformables, motivating a learned policy.

\section{Method}
\textbf{Problem:} A deformable object is a mesh of $N$ vertices on a regular grid
$\mathcal{G}$ of shape $(H,W)$ for a sheet, $(H,1)$ for a linear object, or
$(D,H,W)$ for a volume, with configuration $X\in\mathbb{R}^{N\times3}$ drawn from a
distribution over settled configurations. A touch at vertex $i$ returns a noisy
position $o_i = x_i + \epsilon,\ \epsilon\sim\mathcal{N}(0,\sigma^2 I)$. Given
observations $O_S$ over touched vertices $S$, the estimator $\hat{X}=f(O_S)$
reconstructs the full mesh. We measure the mean per-vertex error
$\mathcal{E}(S)=\frac{1}{N}\sum_i\lVert\hat{x}_i(O_S)-x_i\rVert_2$.

\textbf{Topology-agnostic estimator:} Figure~\ref{fig:pipeline} summarizes the pipeline. Each observation is encoded as a token
concatenating its noisy 3D coordinates with the normalized grid index of the
touched vertex. The grid index has $\dim(\mathcal{G})\in\{1,2,3\}$ components which is the only thing that changes across topologies. Tokens are embedded by an MLP to width $d{=}128$ and mixed by one multi-head self-attention layer so observations
contextualize one another. A fixed bank of \emph{query} tokens, one per output
vertex (features: the vertex's normalized grid coordinates), cross-attends to the
contextualized observations, and a decoder MLP maps each attended query to a 3D
position. Per-vertex cross-attention preserves locality: each output vertex reads the observations individually rather than a single pooled summary, which makes reconstruction sensitive to touch placement. The network is permutation-invariant and accepts any number
of touches. Setting the query grid to $(H,W)$, $(H,1)$, or $(D,H,W)$ yields the cloth, rope, and soft-body estimators from one implementation.

\textbf{Belief and learned acquisition:} We train $M{=}5$ independently initialized
estimators. The ensemble mean is the reconstruction and the per-vertex variance is
an epistemic uncertainty, high where observations under-constrain the shape. We
define an \emph{expected-error-reduction (EER) oracle} that tentatively adds each
candidate touch, reconstructs, and measures the resulting true error. It uses ground truth and serves as an upper bound. A deployable agent instead scores candidates from belief features only (ensemble-mean position, ensemble variance,
grid position, fraction observed, distance to nearest observation) via a small MLP trained by regression against the oracle on on-policy rollouts with $\epsilon$-greedy exploration ($\epsilon{=}0.4$). Greedy-only training instead caused a covariate shift: the policy met unfamiliar belief states at test time and fell below random at the largest budgets. Because features are per candidate vertex, the same acquisition transfers across 1D, 2D, and 3D objects. The learned policy amortizes the oracle's per-candidate lookahead into its weights: at deployment each step needs a single ensemble reconstruction plus a cheap scoring
pass, fewer network evaluations than the oracle.

\section{Experiments}
\textbf{Setup:} We generate configurations in the MuJoCo flex simulator
\cite{mujoco}: a $20{\times}20$ cloth draped over a randomized obstacle into
self-occluded folds, a 40-vertex rope settled into random curves, and a
$6{\times}6{\times}6$ ($216$-vertex) volumetric soft body. We generate $\sim$$10^3$
configurations per object with a held-out test split, train five-member ensembles
($2500$ steps, Adam, $\sigma{=}0.02$), and report mean $\pm$ std over seeds (three
for the estimator, ablation, occlusion, and soft-body results; up to ten for active
sensing). For \emph{estimation} we compare inverse-distance weighting (IDW), a
Gaussian-process implicit-surface (GPIS) reconstruction, and a global-pool
(DeepSets-style \cite{zaheer2017deepsets}) set encoder. For \emph{active sensing} we
compare random, a GP uncertainty baseline \cite{yi16}, an ensemble-variance proxy,
our learned acquisition, and the EER oracle. Unless noted we report percent error
reduction, $100\,(\mathcal{E}_{\mathrm{ref}}-\mathcal{E})/\mathcal{E}_{\mathrm{ref}}$,
against the appropriate reference: random touching for active sensing, and the named
baseline for estimation.

\begin{figure}[h]
\centering
\includegraphics[width=\columnwidth]{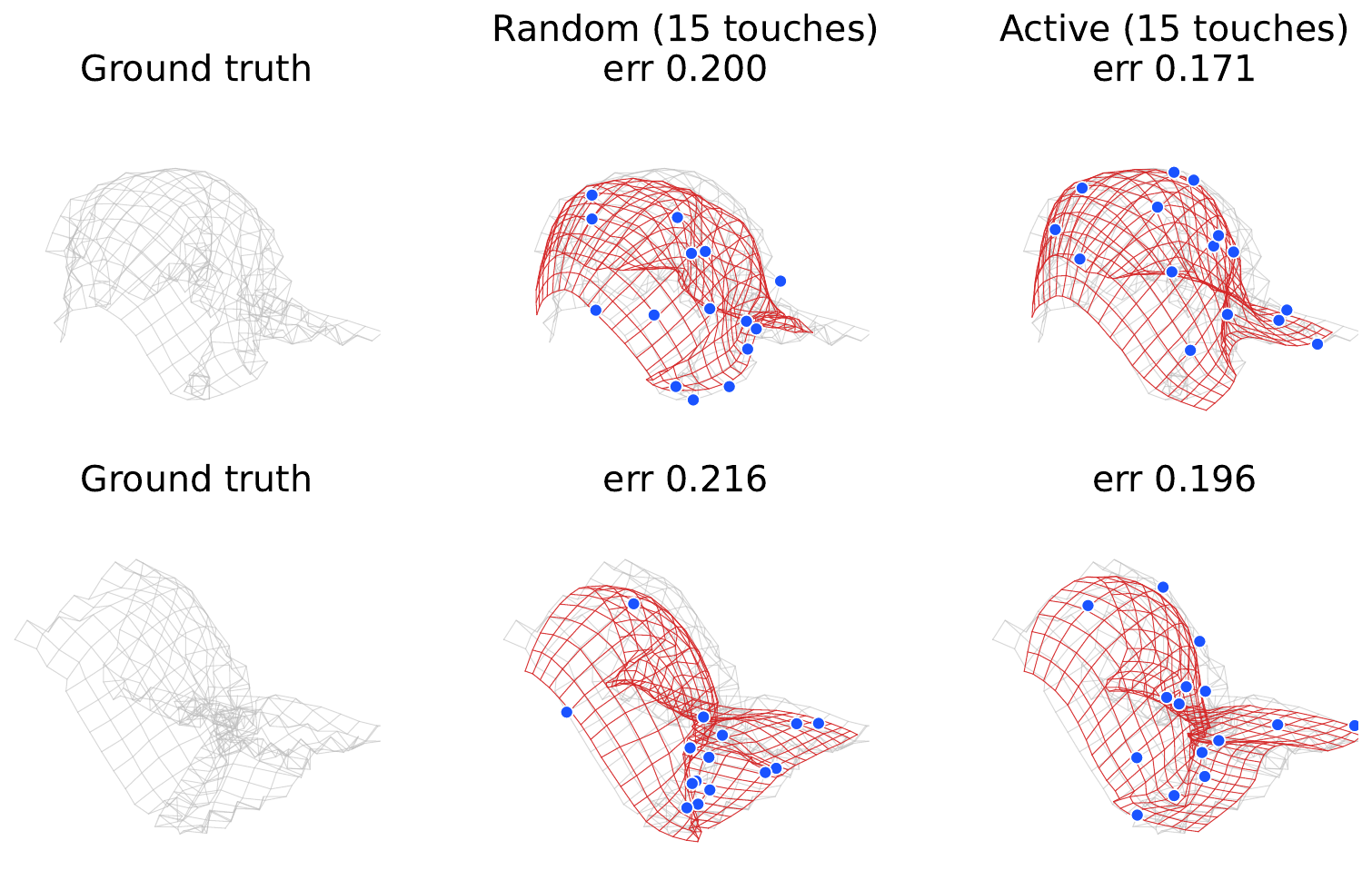}
\caption{Qualitative reconstruction of two heavily-crumpled cloth configurations
from $15$ touches (blue dots). Left: ground truth (grey), middle: reconstruction
from random touches, right: from actively-selected touches, with per-vertex error
annotated. The estimator recovers global fold structure from sparse contacts, and
active selection further reduces error.}
\label{fig:qual}
\end{figure}

\begin{table}[h]
\caption{Estimator error reduction vs.\ non-learned baselines (\%). Budgets
low/mid/high/highest are $K{=}10,15,20,30$ for cloth and $K{=}4,6,8,12$ for rope.}
\label{tab:estim}
\centering
\begin{tabular}{@{}lcccc@{}}
\toprule
& \multicolumn{2}{c}{Cloth} & \multicolumn{2}{c}{Rope} \\
\cmidrule(lr){2-3}\cmidrule(lr){4-5}
Budget & vs.\ GPIS & vs.\ IDW & vs.\ GPIS & vs.\ IDW \\
\midrule
low     & $+74.8$ & $+66.5$ & $+77.9$ & $+66.6$ \\
mid     & $+73.7$ & $+65.1$ & $+77.4$ & $+58.4$ \\
high    & $+71.8$ & $+65.8$ & $+75.3$ & $+53.9$ \\
highest & $+66.6$ & $+65.4$ & $+68.7$ & $+45.3$ \\
\bottomrule
\end{tabular}
\end{table}

The learned estimator outperforms the non-learned baselines by a wide margin.
Table~\ref{tab:estim} reports its error reduction over IDW and GPIS at random
observation sets, and Fig.~\ref{fig:qual} shows example reconstructions. Across
budgets and both object classes, it cuts error by roughly two-thirds. The margin against GPIS is as large as against IDW: a Gaussian-process surface over-smooths the sharp folds of a crumpled mesh, while cross-attention captures them. When we ablate cross-attention against the
global-pool encoder under identical training, cross-attention is consistently better, and the margin grows with budget (from $+4.7\%$ to $+22.4\%$ on cloth as touches accumulate). This says the \emph{locality} of cross-attention, not just the use of a learned set encoder, is what drives reconstruction quality.

The same architecture also handles a volumetric object. With only its query grid
changed to $(6,6,6)$, it reconstructs the soft body from sparse touch and reduces
error over GPIS by $+71$ to $+80\%$, a margin that matches cloth and rope. A single
network thus covers 1D, 2D, and 3D objects.

\begin{figure}[t]
\centering
\includegraphics[width=\columnwidth]{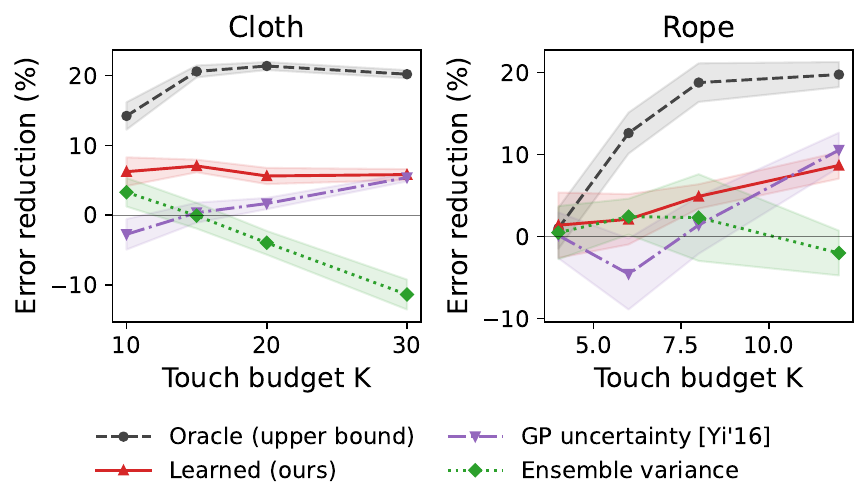}
\caption{Active-sensing error reduction over random vs.\ touch budget, cloth (left)
and rope (right), averaged over seeds (shaded: std). The learned policy beats the
GP baseline at sparse budgets; the ensemble-variance proxy degrades below random;
the oracle bounds the achievable gain.}
\label{fig:active}
\end{figure}

Active sensing gives a real but modest gain, and its size depends on the regime.
Table~\ref{tab:active} and Fig.~\ref{fig:active} report it on cloth and rope. The learned policy reduces error over random by $+5.6$ to $+7.0\%$ across budgets and
beats the GP baseline at sparse budgets. The GP baseline catches up only at the largest budget, where coverage matters more than selection. The ensemble-variance proxy degrades below random as
the budget grows, because maximizing variance concentrates touches on persistently
uncertain outliers rather than informative locations. The oracle reveals substantial headroom ($+14$ to $+21\%$): the gap between deployable and ideal selection, not the absence of headroom, bounds the gains. The same learned acquisition shows the same structure on the 1D rope and on the 3D soft body ($+17$ to $+18\%$ over random at sparse budgets), supporting the
topology-agnostic claim.

\begin{table}[t]
\caption{Active sensing on cloth, \% error reduction vs.\ random.}
\label{tab:active}
\centering
\begin{tabular}{@{}lcccc@{}}
\toprule
$K$ & Variance & GP \cite{yi16} & Learned & Oracle \\
\midrule
10 & $+3.3$  & $-2.7$ & $+6.2$ & $+14.2$ \\
15 & $-0.1$  & $+0.4$ & $+7.0$ & $+20.6$ \\
20 & $-4.0$  & $+1.7$ & $+5.6$ & $+21.4$ \\
30 & $-11.4$ & $+5.4$ & $+5.8$ & $+20.2$ \\
\bottomrule
\end{tabular}
\end{table}

\begin{figure}[t]
\centering
\includegraphics[width=\columnwidth]{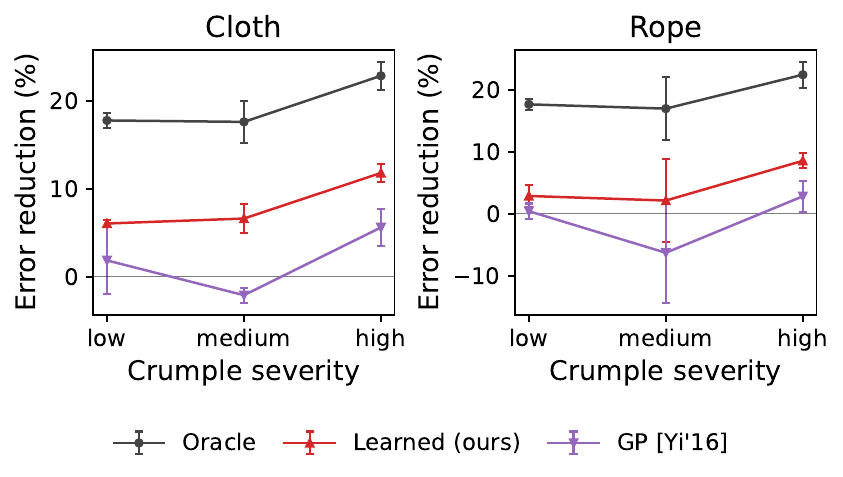}
\caption{The active-sensing advantage grows with self-occlusion: stratifying cloth
(left) and rope (right) configurations by crumple severity, both the learned policy
and the oracle improve most on the most-occluded configurations, the regime where a
vision-based prior is least reliable.}
\label{fig:occlusion}
\end{figure}

The benefit of active sensing grows with occlusion. Stratifying the cloth test set
by crumple severity, the learned policy's error reduction rises from $+6.0\%$ on the
least-occluded third of configurations to $+11.8\%$ on the most-occluded third,
while the oracle headroom rises from $+17.8\%$ to $+22.8\%$ (Fig.~\ref{fig:occlusion}).

A natural question is whether active touch still helps once a camera is available.
We find that it largely does not. With a noisy depth camera and a learned shape
prior alongside touch, active touch lowers occluded-region error by only about $4$
to $6\%$, and the choice of where to touch barely affects the outcome: random and
oracle placement perform almost the same. The camera and the prior already fix most
of the hidden state, so a touch has little left to resolve. Without vision the state
is underdetermined again, and that is where placement begins to matter. 

\section{Discussion}
These results suggest that the leverage in tactile deformable estimation lies in the
estimator, not in where to touch. A locality-aware cross-attention model already
recovers most of what sparse random touches allow across 1D, 2D, and 3D objects.
Active sensing is a real but bounded add-on whose benefit concentrates under
self-occlusion and vanishes once vision fixes the hidden state. Full-mesh estimates
feed directly into deformable planners \cite{diffdyn, gdoom}, though we do not
evaluate planning here. All experiments are in simulation. The position-only touch
model matches a standard arm with descend-until-contact and forward kinematics, so
no specialized tactile skin is required for a hardware demo. The main open algorithmic gap is active selection: closing the oracle-deployable divide, especially under realistic probing, via belief-conditioned policies or amortized lookahead over the belief grid.

\section{Conclusion}
We presented a topology-agnostic estimator that reconstructs full deformable meshes from sparse
touch alone. Learned active sensing helps modestly under occlusion but not when
vision is available. Real-robot validation and stronger active policies are the
next steps.

\bibliographystyle{plainnat}
\bibliography{references}

\end{document}